\documentclass{article}

\PassOptionsToPackage{numbers,compress}{natbib}
\usepackage[preprint]{neurips_2026}

\usepackage{enumitem}    %
\usepackage[utf8]{inputenc} % allow utf-8 input
\usepackage[T1]{fontenc}    % use 8-bit T1 fonts
\usepackage{hyperref}       % hyperlinks
\usepackage{url}            % simple URL typesetting
\usepackage{booktabs}       % professional-quality tables
\usepackage{amsfonts}       % blackboard math symbols
\usepackage{nicefrac}       % compact symbols for 1/2, etc.
\usepackage{microtype}      % microtypography
\usepackage{xcolor}         % colors
\usepackage{amsmath}
\usepackage{amssymb}
\usepackage{amsfonts}
\usepackage{graphicx}
\usepackage{subcaption}
\usepackage{algorithm}
\usepackage{algpseudocode}
\usepackage[table]{xcolor}
\usepackage{graphicx}

\title{DistractMIA: Black-Box Membership Inference on Vision-Language Models via Semantic Distraction}

\author{%
  Hongyi Tang\thanks{Equal contribution.} \\
  The Hong Kong University of Science and Technology \\
  \texttt{htangat@connect.ust.hk} \\
  \And
  Zhihao Zhu\footnotemark[1] \\
  The Hong Kong University of Science and Technology \\
  \texttt{zhihaozhu@ust.hk} \\
  \And
  Yi Yang \\
  The Hong Kong University of Science and Technology \\
  \texttt{imyiyang@ust.hk} \\
}

\date{}

\begin{document}

\maketitle

\begin{abstract}
Vision-language models (VLMs) are trained on large-scale image-text corpora that may contain private, copyrighted, or otherwise sensitive data, motivating membership inference as a tool for training-data auditing. This is especially challenging for deployed VLMs, where auditors typically observe only generated textual responses. Existing VLM membership inference attacks either rely on probability-level signals unavailable in such settings, or use mask-based semantic prediction tasks whose effectiveness depends on object-centric visual assumptions.
To address these limitations, we propose DistractMIA, an output-only black-box framework based on semantic distraction. Rather than removing visual evidence, DistractMIA preserves the original image, inserts a known semantic distractor, and measures how generated responses change. This design is motivated by the intuition that member samples remain more anchored to the original image semantics, while non-member samples are more easily redirected toward the distractor. To make this signal reliable, DistractMIA calibrates distractor configurations on a reference set and derives membership scores from repeated textual generations, capturing response stability and distractor uptake without accessing logits, probabilities, or hidden states. Experiments across multiple VLMs and benchmarks show that DistractMIA consistently outperforms both output-only and stronger-access baselines. Its performance on a medical benchmark further demonstrates applicability beyond object-centric natural images.
\end{abstract}

\section{Introduction}

\noindent Recent advances in vision-language models (VLMs) have enabled strong performance in tasks such as visual question answering, image captioning, and multimodal reasoning~\citep{hu2022scaling, chappuis2022prompt, wu2024visionllm}. These capabilities are largely driven by training on massive image-text corpora collected from diverse sources~\citep{zhang2024rs5m, gao2020pile}. However, such data may contain private photographs, medical images, copyrighted content, or user-uploaded data, raising serious concerns about whether sensitive or proprietary images have been memorized by deployed models~\citep{getty2025stability, wakabayashi2019google}. This creates a growing need for privacy auditing tools that can assess the exposure of individual training examples. Membership inference attacks (MIAs), which aim to determine whether a specific data sample was included in the training dataset of a target model~\citep{shokri2017membership, carlini2022membership}, provide a natural framework for auditing training-data privacy risks in VLMs.

\noindent However, existing MIAs are often mismatched with realistic VLM auditing scenarios. Many strong attacks rely on likelihood-based or internal signals, such as token probabilities, losses, or hidden representations~\citep{shidetecting, tang2025identifying}. These assumptions are difficult to satisfy for deployed VLM services, where users typically observe only textual responses to image-text queries. Recent output-only VLM attacks, such as KCMP, move closer to this setting by constructing mask-based semantic prediction tasks and inferring membership from whether the model can recover masked object or color information~\citep{yin2025black}.

\noindent Despite this progress, mask-based semantic prediction still faces two key limitations. A correct prediction after masking does not necessarily indicate memorization: the model may recover the answer from residual visual cues, surrounding context, or general visual-language priors. Moreover, its effectiveness depends on the availability of clearly identifiable semantic entities, the quality of the mask, and the design of the prediction task. This can limit its applicability to images with low semantic diversity or weakly separable structures, such as medical scans or document images, where constructing informative mask-based probes becomes substantially more difficult~\citep{yin2025black, xie2024rethinking}.

\noindent To address these limitations, we propose DistractMIA, a output-only semantic distraction framework for black-box VLM membership inference. Instead of removing visual evidence and relying on reconstruction accuracy, DistractMIA preserves the original image and introduces an additional, known semantic distractor. This shifts the probe from semantic completion to semantic distraction. The key intuition is that membership can be reflected in how strongly the model’s generation remains tied to the original image when a distracting semantic cue is introduced. Since the distractor is externally introduced, our method does not rely on the presence of clearly segmentable objects in the original image, making it applicable beyond natural images with rich object semantics.

\noindent This semantic-distraction design raises two key challenges. First, the effect of a distractor depends on its visual form, location, and size. A weak distractor may fail to induce observable response changes, while an overly dominant one may overwhelm the original image for both members and non-members. Second, output-only access requires membership evidence to be extracted solely from generated text. Since VLM responses are stochastic and language-mediated, the method needs to convert changes in textual behavior into reliable membership features. 
To address these challenges, DistractMIA first uses an independent calibration set to automatically search for distractor configurations that better separate member and non-member behavior, and then freezes the selected configuration for evaluation. With this configuration fixed, it performs repeated queries on both the original and distracted images, allowing the attack to estimate stable text-level behavioral patterns rather than relying on a single response. From these responses, DistractMIA extracts output-only behavioral features, including response stability and distraction-induced changes, and aggregates them into a membership score for the target image. The attack relies solely on generated text and does not require logits, token probabilities, or internal representations.

% \noindent \zzh{We evaluate DistractMIA on three representative VLMs, including LLaVA-1.5, MiniGPT-4, and LLaMA2-Accessory, using the VL-MIA/Flickr and VL-MIA/DALL$\cdot$E benchmarks. DistractMIA consistently outperforms output-only baselines and remains competitive with stronger probability-access methods. For example, on VL-MIA/Flickr, DistractMIA improves the average AUROC from $0.525$ for the strongest output-only baseline to $0.675$ at the single-sample level. It also benefits substantially from set-level aggregation, suggesting that the distraction-induced signal is stable across samples. ROC analysis further shows that this advantage is not tied to a particular decision threshold. Attention-shift analysis supports the proposed mechanism: non-member samples exhibit larger shifts toward inserted distractors, while member samples remain more anchored to the original image semantics. Ablation and sensitivity analyses show that DistractMIA remains robust under different distractor configurations, scoring variants, and query budgets. We further evaluate DistractMIA on a medical-domain PMC benchmark with LLaVA-Med, where object-centric probing is difficult to apply, and find that semantic distraction remains effective for images with weaker object boundaries.}

We evaluate DistractMIA on three representative VLMs, including LLaVA-1.5~\citep{liu2023visual}, MiniGPT-4~\citep{zhu2023minigpt}, and LLaMA2-Accessory~\citep{touvron2023llama}, using the VL-MIA/Flickr and VL-MIA/DALL$\cdot$E benchmarks~\citep{li2024membership}. DistractMIA consistently outperforms output-only baselines and also surpasses stronger probability-access methods. For example, on VL-MIA/Flickr, DistractMIA improves the average AUROC from $0.556$ for the strongest output-only baseline to $0.916$ under $K=10$ set-level evaluation. This gain suggests that the distraction-induced membership signal is stable across samples and can be amplified through aggregation. To better understand this signal, we analyze attention shifts under semantic distraction and find that non-member samples exhibit larger shifts toward inserted distractors, while member samples remain more anchored to the original image semantics. We further evaluate DistractMIA on a medical-domain benchmark, where object-centric probing is difficult to apply, and find that semantic distraction remains effective for images with weaker object boundaries.

The contributions of this paper are three-fold:
\begin{itemize}[leftmargin=1.2em, itemsep=1pt, topsep=1pt, parsep=1pt, partopsep=1pt]

\item We identify distractor-induced response shift as an output-observable membership signal for VLMs, showing that membership can be inferred from how generated responses change after inserting a known semantic distractor.

\item We develop DistractMIA, a calibrated output-only framework for translating semantic-distraction behavior into membership scores. It uses automatic distractor configuration search to select reliable distractions, and text-level scoring to extract stable membership evidence from stochastic VLM generations.

\item We evaluate DistractMIA across multiple VLMs, benchmarks, and a medical-domain setting, demonstrating strong performance across diverse evaluation scenarios and consistently outperforming both output-only and stronger-access baselines.

\end{itemize}

\section{Related Work}

Membership inference attacks (MIAs) aim to determine whether a given sample was included in the training data of a target model~\cite{hu2022membership}. Existing methods mainly differ in how they extract membership signals from model behavior~\cite{salem2018ml}. In vision-language models (VLMs), this problem becomes particularly challenging because deployed systems usually expose only generated text, while the outputs themselves can vary substantially across repeated queries~\cite{li2024membership}. In this work, we focus on the practical black-box setting~\cite{yin2025black}, where the attacker can query the target VLM but cannot access model parameters, gradients, logits, or token probabilities.

\subsection{White-box and Gray-box Membership Inference}

Early MIA methods were primarily studied under white-box or gray-box assumptions, where the attacker has access to internal model statistics such as prediction confidence, loss values, likelihood scores, hidden states, or gradients~\cite{shokri2017membership,salem2018ml,ibanez2025lumia}. These approaches were first developed for classification models and were later extended to large language models (LLMs) and VLMs.

In LLMs, many methods rely on token-level statistics such as likelihood, entropy, or Min-K\% to distinguish member from non-member samples~\cite{duan2024membership,li2024membership,shi2023detecting}. Under stronger white-box assumptions, prior work has also explored hidden-state similarity, intermediate representations, and attention activations as indicators of memorization behavior~\cite{wu2026image,tang2025identifying,makhija2026neural}. More recently, similar ideas have been adapted to VLMs, including methods based on R\'enyi entropy, token probability distributions, and embedding similarity~\cite{li2024membership,pu2024embedding}.

A common limitation of these methods is their dependence on structured model outputs and relatively stable output distributions. In practical VLM deployments, however, commercial systems often expose only free-form textual responses rather than logits or token probabilities. In addition, the stochastic nature of text generation can lead to substantially different outputs for the same image across repeated queries, which weakens the reliability of probability-based membership signals.

\subsection{Black-box Membership Inference for Vision-Language Models}

To reduce dependence on internal model access, recent work has explored black-box membership inference based on input perturbations, such as Gaussian noise, color jitter, random cropping, and data augmentation~\citep{song2021systematic,kaya2021does}. Membership signals are constructed by observing how model outputs respond to small input changes; however, these low-level visual perturbations do not substantially alter image semantics, making the resulting signals weak and unstable.

Other approaches perform stronger semantic-level interventions, for example by masking image regions and asking the model to infer the missing content~\citep{yin2025black}. While these methods can amplify memorization effects, their effectiveness depends on whether the remaining image still preserves sufficient semantic information. In low-semantic-density domains such as medical images, masking important regions can easily produce generic outputs, making membership signals unreliable~\citep{yin2025black, xie2024rethinking}. % We refer to this phenomenon as \emph{semantic collapse}.

% DistractMIA differs from existing methods in how the membership signal is constructed. Rather than relying on probability-based statistics or removing information from the image, our method inserts small semantic distractors and observes how the generated description responds. The resulting signal emerges from semantic competition between the original image content and the inserted distractor. Empirically, non-member samples are more easily redirected toward the distractor, while member samples tend to remain semantically anchored to the original visual content.

DistractMIA differs from existing methods in how the membership signal is constructed. Rather than relying on probability-based statistics or removing information from the image, our method inserts small semantic distractors and measures output-level changes in the generated descriptions. This makes DistractMIA complementary to existing output-only VLM MIAs, which mainly rely on prediction correctness under constructed visual tasks.

\section{Methodology}

\begin{figure}[t!]
    \centering
    \includegraphics[width=\textwidth]{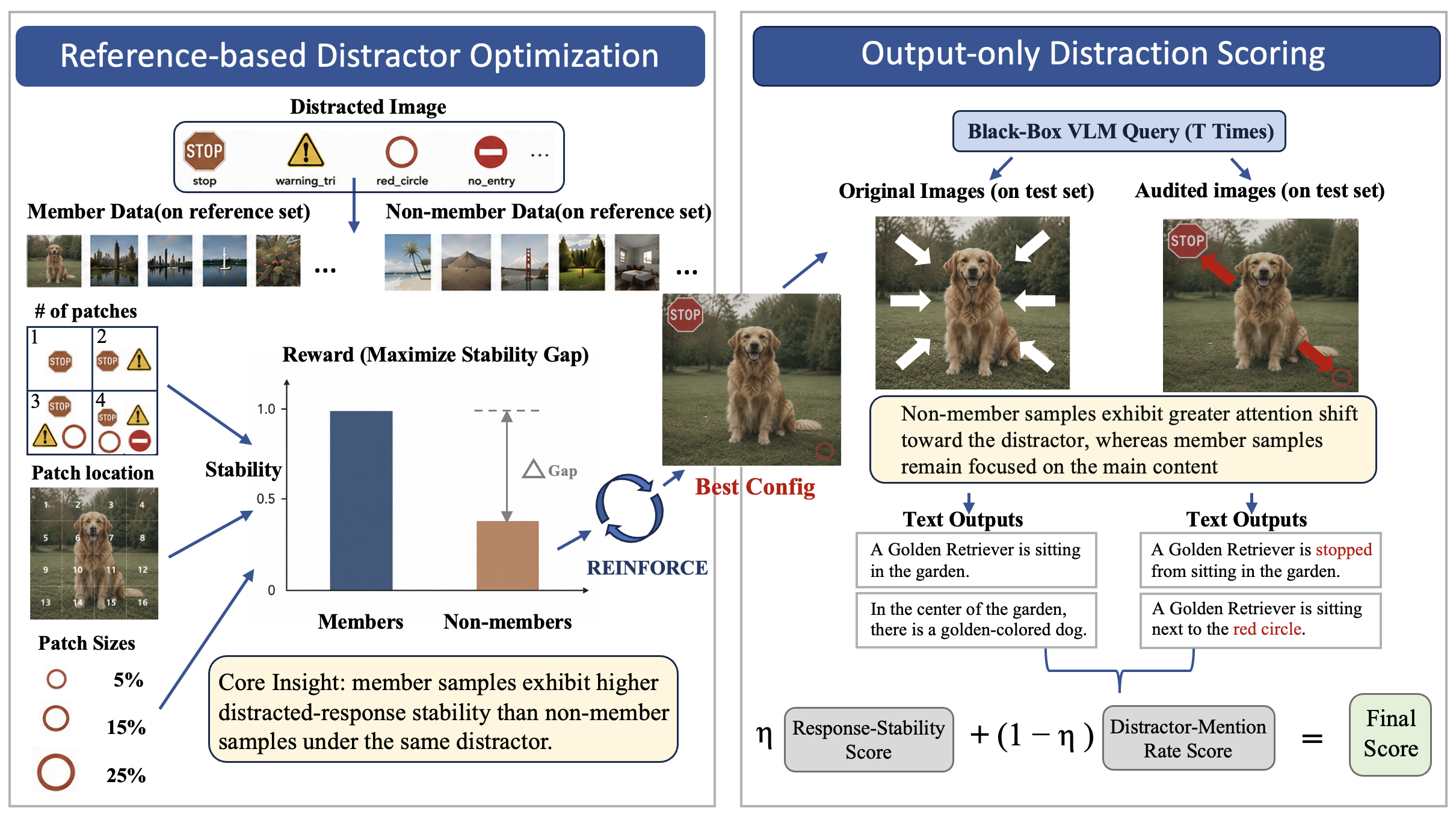}
    \caption{Overview of DistractMIA. Semantic distractors are applied to images to evaluate VLM output stability for membership inference.}
    \label{fig:distractmia_method}
\end{figure}

\subsection{Problem Formulation}

Given a target vision-language model $M$ and an audited image $x$, image-level membership inference aims to determine whether $x$ was included in the training data of $M$. The adversary queries the model with $x$ and a prompt $q$, and observes the generated response $y \sim M(x,q)$. The attack then outputs a membership score $s(M,x)$, where a larger score indicates stronger evidence of membership.

We consider a calibrated output-only black-box setting. The adversary can query the target VLM and observe only generated text, without access to model parameters, gradients, logits, or hidden representations. Following prior work that uses reference data for membership calibration~\citep{tang2025identifying, zhang2024pretraining}, we assume access to a small reference set from a public and easily obtainable source, such as COCO~\citep{lin2014microsoft}. The reference set is used only to select distractor configurations and is disjoint from the final evaluation set. For many deployed or publicly available models, the use of common public datasets is documented, making such datasets suitable reference sources for calibration~\citep{zhuminigpt, wang2024qwen2}.

\subsection{Overview of DistractMIA}

DistractMIA infers membership based on the stability of VLM generations under semantic distraction. Prior studies have shown that large models can memorize training examples, leading to more stable or over-confident behavior on seen data~\citep{carlini2022quantifying, schwarzschild2024rethinking}. In VLMs, such memorization may manifest as stronger anchoring to the original image semantics. Motivated by this observation, DistractMIA inserts an external semantic distractor while preserving the original image content. Member images tend to remain more stable under distraction, whereas non-member images are more likely to exhibit larger changes in their generated descriptions. Figure~\ref{fig:distractmia_method} provides an overview of the framework.

The method has two stages. In the calibration stage, DistractMIA searches for a distractor configuration that exposes a consistent generation-stability gap between member and non-member samples in a small reference set. The selected configuration is fixed after this stage. In the inference stage, DistractMIA applies the selected distractor to each audited image and queries the target VLM multiple times on both the original and distracted images. It then computes a membership score from output-level textual signals capturing generation stability and distraction-induced shifts.

\subsection{Reference-based Distractor Optimization}

We model a distractor configuration as $\mathcal{P}=\{p_1,\ldots,p_L\}$, where $L$ is the number of inserted patches. Each patch $p_i=(\ell_i, a_i, r_i)$ specifies its location, semantic pattern, and relative size, respectively. Given an image $x$, applying $\mathcal{P}$ produces a distracted image
% \[
$\tilde{x}=\mathcal{T}(x,\mathcal{P}).$
%\]
The goal of the calibration stage is to identify a configuration $\mathcal{P}$ that yields a clear member--non-member separability under distraction, measured through generation stability on the reference set.

For optimization, we use a low-variance surrogate score based only on the distracted image. Given $\tilde{x}$, we query the target model $T$ times and obtain responses $\{\tilde{y}_1,\ldots,\tilde{y}_T\}$. Each response is converted into a normalized word set $\tilde{W}_i$ by lowercasing, splitting, and removing common stop words. We define the distracted-response stability as
\[
S_{\mathrm{ref}}(x;\mathcal{P})
=
\frac{1}{\binom{T}{2}}
\sum_{i<j}
\frac{|\tilde{W}_i\cap \tilde{W}_j|}
{|\tilde{W}_i\cup \tilde{W}_j|}.
\]
This score measures how consistently the model describes the distracted image across stochastic generations. We use it as a surrogate for distractor selection because it provides a continuous signal and aggregates pairwise similarities across multiple generations, yielding lower variance than sparse decision-based objectives.

Let the reference set contain known member samples $\mathcal{R}_m$ and known non-member samples $\mathcal{R}_n$. To reduce overfitting to a small reference set, we further partition both $\mathcal{R}_m$ and $\mathcal{R}_n$ into two disjoint subsets, yielding two stratified reference splits:
$\mathcal{R}^{(1)}=(\mathcal{R}^{(1)}_m,\mathcal{R}^{(1)}_n)$ and
$\mathcal{R}^{(2)}=(\mathcal{R}^{(2)}_m,\mathcal{R}^{(2)}_n)$.
Here, the subscripts $m$ and $n$ indicate membership labels, while the superscripts index the two reference splits. For each split, we compute the member--non-member separation:
\[
\Delta_r(\mathcal{P})
=
\mathbb{E}_{x\in \mathcal{R}^{(r)}_m}
\left[
S_{\mathrm{ref}}(x;\mathcal{P})
\right]
-
\mathbb{E}_{x\in \mathcal{R}^{(r)}_n}
\left[
S_{\mathrm{ref}}(x;\mathcal{P})
\right],
\quad r\in\{1,2\}.
\]
A positive value of $\Delta_r(\mathcal{P})$ indicates that member samples exhibit higher response stability than non-member samples under the same distractor. We therefore define the distractor-selection reward as
\[
R(\mathcal{P})
=
\alpha \max(0,\Delta_1(\mathcal{P}))
+
\beta \max(0,\Delta_2(\mathcal{P}))
-
\lambda |\Delta_1(\mathcal{P})-\Delta_2(\mathcal{P})|.
\]
The first two terms reward configurations that produce positive member--non-member separation on both reference splits. The last term penalizes inconsistent separation across splits, discouraging configurations that only work for a small subset of reference samples. % This objective favors distractors that induce a stable and reproducible separation in generation stability.

% We optimize this objective over a discrete distractor search space using a policy $\pi_\theta(\mathcal{P})$. The policy samples candidate configurations and is updated with a REINFORCE-style policy gradient:
% \[
% \nabla_\theta J(\theta)
% =
% \mathbb{E}_{\mathcal{P}\sim\pi_\theta}
% \left[
% (R(\mathcal{P})-b_t)
% \nabla_\theta \log \pi_\theta(\mathcal{P})
% \right].
% \]
% Here, $b_t$ is an exponential moving average of previous rewards, which serves as a standard variance-reduction baseline for policy-gradient optimization~\citep{williams1992simple}. After optimization, we select the configuration with the highest reward on the reference set and fix it for all subsequent inference.

We optimize the distractor configuration over a discrete search space. Specifically, we parameterize a policy $\pi_\theta$ that assigns a probability to each candidate configuration $\mathcal{P}$, and define the objective as the expected distractor-selection reward:
\[
J(\theta)=\mathbb{E}_{\mathcal{P}\sim\pi_\theta}\left[R(\mathcal{P})\right].
\]
The reward $R(\mathcal{P})$ is obtained by querying the target model on the reference set, and is not differentiable with respect to the discrete configuration choices. We therefore optimize the policy parameters using the REINFORCE gradient estimator:
\[
\nabla_\theta J(\theta)
=
\mathbb{E}_{\mathcal{P}\sim\pi_\theta}
\left[
(R(\mathcal{P})-b_t)
\nabla_\theta \log \pi_\theta(\mathcal{P})
\right].
\]
Here, $\log \pi_\theta(\mathcal{P})$ denotes the log-probability of the sampled configuration, and $b_t$ is an exponential moving average of previous rewards used to reduce the variance of the gradient estimator~\citep{williams1992simple}.
After optimization, we select the configuration with the highest reward on the reference set and fix it for all subsequent inference.

\subsection{Output-only Distraction Scoring}

With the optimized distractor configuration $\mathcal{P}^*$ fixed, DistractMIA computes membership scores for audited images without further optimization. For each image $x$, we apply $\mathcal{P}^*$ to obtain the distracted image $\tilde{x}$. DistractMIA then compares the target model's textual behavior on $x$ and $\tilde{x}$, so that each image serves as its own baseline.

For behavioral scoring, each response is converted into a normalized word set. For any input image $z$, we query the target model $T$ times and obtain responses $\{y_1(z),\ldots,y_T(z)\}$, with corresponding word sets $\{W_1(z),\ldots,W_T(z)\}$. We define the response-stability score as
\[
S(z)
=
\frac{1}{\binom{T}{2}}
\sum_{i<j}
\frac{|W_i(z)\cap W_j(z)|}
{|W_i(z)\cup W_j(z)|}.
\]
This score measures how consistently the model generates across stochastic generations.

However, stability alone does not capture whether the inserted distractor explicitly appears in the generated text. To capture this effect, let $\mathcal{K}_{\mathrm{dist}}$ denote the keyword set associated with the distractor. We define the distractor-mention rate as
\[
M(z)
=
\frac{1}{T}
\sum_{i=1}^{T}
\mathbf{1}
\left[
W_i(z) \cap \mathcal{K}_{\mathrm{dist}}\neq \emptyset
\right].
\]
A larger mention rate indicates stronger uptake of distractor-related semantics in the responses.

We combine these two quantities into a distraction-response score:
\[
G(z)=\eta S(z)-(1-\eta)M(z),
\]
where $\eta$ is fixed before evaluation. The stability term rewards consistent generations, while the mention term penalizes explicit distractor uptake. Finally, DistractMIA computes the membership score as the change in this response score induced by the distractor:
\[
A(x)=G(\tilde{x})-G(x).
\]
Using $G(x)$ as a sample-specific baseline accounts for the intrinsic generation stability of the original image before distraction. A larger $A(x)$ indicates that the model remains more stable and less distracted after the semantic intervention, providing stronger evidence of membership. All quantities in $A(x)$ are computed solely from the output of the target model.

\section{Experiments}

\subsection{Experimental Settings}
We first present the experimental setup, including the evaluation datasets, target models, baseline methods, evaluation metrics, and implementation details.

\paragraph{Datasets.} % \footnote{Dataset construction details follow the VL-MIA benchmark protocol~\citep{li2024membership}.}
We evaluate DistractMIA on two standard VL-MIA benchmarks, Flickr and DALL$\cdot$E, using the official member/non-member splits~\citep{li2024membership}. Both benchmarks are balanced. VL-MIA/Flickr follows a temporal split, where non-member images are collected after the release of the target models to reduce potential training overlap. VL-MIA/DALL$\cdot$E provides a more challenging paired setting, where each non-member image is generated to be semantically aligned with a corresponding member image, reducing the effect of temporal distribution shift~\citep{duanmembership}. In addition, we evaluate on a medical-domain PMC benchmark~\citep{lin2023pmc} with LLaVA-Med~\citep{li2023llava}, testing DistractMIA on low-semantic-density images where object-centric mask probes are difficult to construct.

\paragraph{Target models.}
We evaluate on multiple representative open-source vision-language models: LLaVA-1.5~\citep{liu2023visual}, MiniGPT-4~\citep{zhu2023minigpt}, and LLaMA2-Accessory~\citep{touvron2023llama}. These models differ in architecture, pretraining data, and instruction-tuning procedures, allowing us to evaluate whether the proposed distraction-based signal remains consistent across different VLM instantiations. For DistractMIA, each target model is queried under an output-only black-box setting. The attack observes only generated textual responses and does not access internal states, logits, or token-level probabilities.

\paragraph{Baselines.}
We compare DistractMIA with existing black-box VLM membership inference baselines. KCMP~\citep{yin2025black} is the most closely related method: it constructs masked semantic prediction tasks and infers membership from whether the target VLM can recover the masked semantic information. We also include Image Infer~\citep{song2018contextual}, which infers membership from caption-level similarity across repeated queries.
Since existing output-only baselines for VLM membership inference are limited, we further report a set of stronger-access baselines that use token-level probabilities or output distributions. These include Perplexity~\citep{horgan1995complexity}, Aug-KL~\citep{liu2021encodermi}, Min-K\% Prob~\citep{shi2023detecting}, and Max-Prob-Gap, ModR\'enyi, and MaxR\'enyi-K\%~\citep{li2024membership}. These methods are not available in a strict output-only API setting, but provide useful references for comparing DistractMIA against attacks with access to richer model signals.

\paragraph{Evaluation metrics.}
We use the Area Under the ROC Curve (AUROC) as the evaluation metric. For sample-level inference, AUROC is computed directly from individual membership scores. For set-level inference, we first average the membership scores of $K$ samples within each set and then compute AUROC over the resulting set-level scores. We report results for $K \in \{1,10,30\}$.

\paragraph{Implementation details.}
We use a COCO reference set~\citep{lin2014microsoft} for distractor configuration search, containing 50 member samples and 50 non-member samples. The reference set is used only during calibration and is disjoint from the final evaluation sets. For the distractor-selection objective, we set $\alpha=1$, $\beta=0.5$, and $\lambda=0.2$. For output-only scoring, we set $\eta=0.7$ to assign higher weight to generation stability while still penalizing explicit distractor uptake. To estimate distraction signals, We use $T=5$ stochastic generations per input unless otherwise specified.

\begin{table*}[t!]
\centering
\scriptsize
\setlength{\tabcolsep}{2.4pt}
\renewcommand{\arraystretch}{1.12}
\resizebox{0.95\textwidth}{!}{
\begin{tabular}{c|ccc|ccc|ccc|ccc}
\toprule
\multicolumn{13}{c}{\textbf{VL-MIA/Flickr}} \\
\midrule
Method
& \multicolumn{3}{c|}{LLaVA}
& \multicolumn{3}{c|}{LLaMA2-A}
& \multicolumn{3}{c|}{MiniGPT-4}
& \multicolumn{3}{c}{Average} \\
\cmidrule(lr){2-4} \cmidrule(lr){5-7} \cmidrule(lr){8-10} \cmidrule(lr){11-13}
& K=1 & K=10 & K=30
& K=1 & K=10 & K=30
& K=1 & K=10 & K=30
& K=1 & K=10 & K=30 \\
\midrule

Perplexity
& 0.637 & 0.898 & 0.968
& 0.617 & 0.794 & 0.897
& 0.634 & 0.842 & 0.961
& 0.629 & 0.845 & 0.942 \\

Aug-KL
& 0.516 & 0.576 & 0.565
& 0.524 & 0.527 & 0.618
& 0.494 & 0.476 & 0.515
& 0.511 & 0.526 & 0.566 \\

Max-Prob-Gap
& 0.638 & 0.787 & 0.892
& 0.614 & 0.820 & 0.937
& 0.596 & 0.778 & 0.903
& 0.616 & 0.795 & 0.911 \\

Min-K\%
& 0.645 & 0.873 & 0.897
& 0.568 & 0.743 & 0.886
& 0.601 & 0.717 & 0.877
& 0.605 & 0.778 & 0.887 \\

ModRényi
& 0.559 & 0.797 & 0.895
& 0.511 & 0.697 & 0.862
& 0.539 & 0.752 & 0.870
& 0.536 & 0.749 & 0.876 \\

MaxRényi-K\%
& 0.583 & 0.763 & 0.896
& 0.578 & 0.750 & 0.817
& 0.580 & 0.751 & 0.821
& 0.580 & 0.754 & 0.845 \\

\midrule
\rowcolor{gray!15}
Image Infer (Rouge)
& 0.516 & 0.524 & 0.529
& 0.508 & 0.537 & 0.568
& 0.523 & 0.546 & 0.572
& 0.516 & 0.536 & 0.556 \\

\rowcolor{gray!15}
Image Infer (MPNet)
& 0.502 & 0.517 & 0.506
& 0.491 & 0.505 & 0.523
& 0.487 & 0.521 & 0.464
& 0.493 & 0.514 & 0.497 \\

\rowcolor{gray!15}
KCMP
& 0.532 & 0.580 & 0.639
& 0.503 & 0.525 & 0.525
& 0.544 & 0.562 & 0.618
& 0.526 & 0.556 & 0.594 \\

\midrule
\rowcolor{gray!15}
\textbf{DistractMIA}
& \textbf{0.671} & \textbf{0.911} & \textbf{0.989}
& \textbf{0.711} & \textbf{0.961} & \textbf{0.999}
& \textbf{0.643} & \textbf{0.875} & \textbf{0.980}
& \textbf{0.675} & \textbf{0.916} & \textbf{0.989} \\

\bottomrule
\end{tabular}
}
\caption{VL-MIA/Flickr Results.}
\label{tab:flickr_results}
\end{table*}

\begin{table*}[t!]
\centering
\scriptsize
\setlength{\tabcolsep}{2.4pt}
\renewcommand{\arraystretch}{1.12}
\resizebox{0.95\textwidth}{!}{
\begin{tabular}{c|ccc|ccc|ccc|ccc}
\toprule
\multicolumn{13}{c}{\textbf{VL-MIA/DALL$\cdot$E}} \\
\midrule
 Method
& \multicolumn{3}{c|}{LLaVA}
& \multicolumn{3}{c|}{LLaMA2-A}
& \multicolumn{3}{c|}{MiniGPT-4}
& \multicolumn{3}{c}{Average} \\
\cmidrule(lr){2-4} \cmidrule(lr){5-7} \cmidrule(lr){8-10} \cmidrule(lr){11-13}
& K=1 & K=10 & K=30
& K=1 & K=10 & K=30
& K=1 & K=10 & K=30
& K=1 & K=10 & K=30 \\
\midrule

Perplexity
& 0.558 & 0.742 & 0.876
& 0.506 & 0.532 & 0.567
& 0.525 & 0.633 & 0.681
& 0.530 & 0.635 & 0.708 \\

Aug-KL
& 0.522 & 0.525 & 0.604
& 0.513 & 0.574 & 0.679
& 0.534 & 0.640 & 0.628
& 0.523 & 0.580 & 0.637 \\

Max-Prob-Gap
& 0.569 & 0.745 & 0.865
& 0.527 & 0.603 & 0.674
& 0.537 & 0.569 & 0.628
& 0.544 & 0.639 & 0.722 \\

Min-K\%
& 0.558 & 0.673 & 0.777
& 0.513 & 0.528 & 0.569
& 0.527 & 0.545 & 0.589
& 0.533 & 0.582 & 0.645 \\

ModRényi
& 0.540 & 0.649 & 0.789
& 0.510 & 0.536 & 0.607
& 0.539 & 0.612 & 0.640
& 0.530 & 0.599 & 0.679 \\

MaxRényi-K\%
& 0.548 & 0.660 & 0.764
& 0.528 & 0.574 & 0.675
& 0.525 & 0.640 & 0.657
& 0.534 & 0.625 & 0.699 \\

\midrule
\rowcolor{gray!15}
Image Infer (Rouge)
& 0.497 & 0.506 & 0.521
& 0.502 & 0.536 & 0.538
& 0.521 & 0.510 & 0.543
& 0.507 & 0.517 & 0.534 \\

\rowcolor{gray!15}
Image Infer (MPNet)
& 0.493 & 0.503 & 0.515
& 0.519 & 0.522 & 0.491
& 0.502 & 0.511 & 0.526
& 0.505 & 0.512 & 0.511 \\

\rowcolor{gray!15}
KCMP
& 0.501 & 0.524 & 0.522
& 0.518 & 0.527 & 0.541
& 0.503 & 0.554 & 0.563
& 0.507 & 0.535 & 0.542 \\

\midrule
\rowcolor{gray!15}
\textbf{DistractMIA}
& \textbf{0.582} & \textbf{0.753} & \textbf{0.891}
& \textbf{0.530} & \textbf{0.606} & \textbf{0.683}
& \textbf{0.542} & \textbf{0.646} & \textbf{0.684}
& \textbf{0.551} & \textbf{0.668} & \textbf{0.753} \\

\bottomrule
\end{tabular}
}
\caption{VL-MIA/DALL-E Results.}
\label{tab:dalle_results}
\end{table*}

\subsection{Main Results}

Table~\ref{tab:flickr_results} and Table~\ref{tab:dalle_results} report the main membership inference results on VL-MIA/Flickr and VL-MIA/DALL$\cdot$E. DistractMIA achieves the strongest performance among output-only methods on both benchmarks. The improvement is already visible at the single-sample level: for example, on VL-MIA/Flickr, DistractMIA obtains an average AUROC of $0.675$ at $K=1$, compared with $0.525$ for KCMP and $0.516$ for Image Infer (Rouge). This suggests that the response change induced by semantic distraction provides a more discriminative output-only membership signal than existing generation-consistency or mask-based probes.

Set-level aggregation further improves DistractMIA. On VL-MIA/Flickr, its average AUROC increases from $0.675$ at $K=1$ to $0.989$ at $K=30$, while on VL-MIA/DALL$\cdot$E it increases from $0.551$ to $0.753$. In contrast, output-only baselines show smaller gains under the same setting. For example, KCMP improves only from $0.526$ to $0.594$ on Flickr and from $0.507$ to $0.542$ on DALL$\cdot$E. The absolute performance on VL-MIA/DALL$\cdot$E is lower than on Flickr, as DALL$\cdot$E uses semantically aligned non-member counterparts and weakens coarse visual cues. Even under this harder setting, DistractMIA achieves the strongest membership inference performance.

\begin{figure}[t!]
    \centering
    \includegraphics[width=0.9\linewidth]{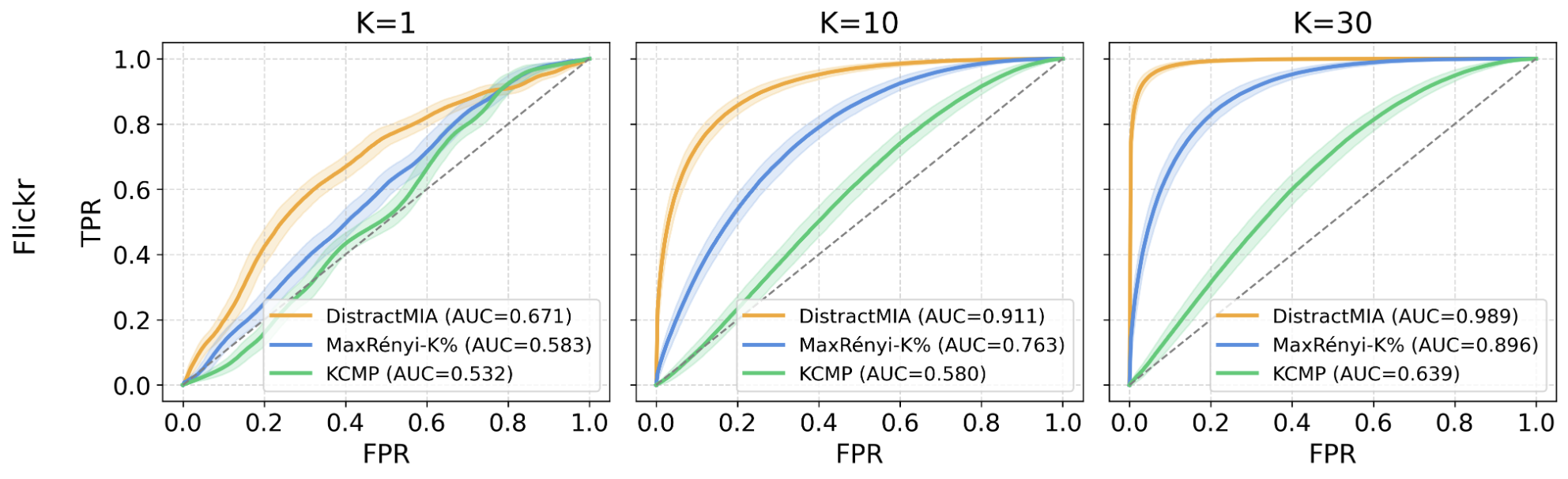}
    \caption{
    ROC curves on LLaVA for VL-MIA/Flickr under different set sizes $K$.
    }
    \label{fig:roc}
\end{figure}

To further complement the tabular results, Figure~\ref{fig:roc} plots ROC curves on LLaVA for representative methods under different set sizes. The curves follow the same trend as Tables~\ref{tab:flickr_results} and~\ref{tab:dalle_results}, showing that DistractMIA provides stronger separation between members and non-members across decision thresholds. This confirms that the improvement is not tied to a particular threshold choice.

% \begin{figure}[t]
%     \centering

%     % ===== Row 1: Flickr =====
%     \begin{subfigure}{\linewidth}
%         \centering
%         \includegraphics[width=\linewidth]{figs/roc_llava_flickr.png}
%         \caption{VL-MIA/Flickr}
%     \end{subfigure}

%     \vspace{6pt}

%     % ===== Row 2: DALL·E =====
%     \begin{subfigure}{\linewidth}
%         \centering
%         \includegraphics[width=\linewidth]{figs/roc_llava_dalle.png}
%         \caption{VL-MIA/DALL$\cdot$E}
%     \end{subfigure}

%     \caption{
%     ROC curves on LLaVA under different set sizes $K$.
%     DistractMIA consistently outperforms MaxR\'enyi-K\% and KCMP across both datasets.
%     }
%     \label{fig:roc}
% \end{figure}

\begin{figure}[t!]
    \centering
    \includegraphics[width=0.8\linewidth]{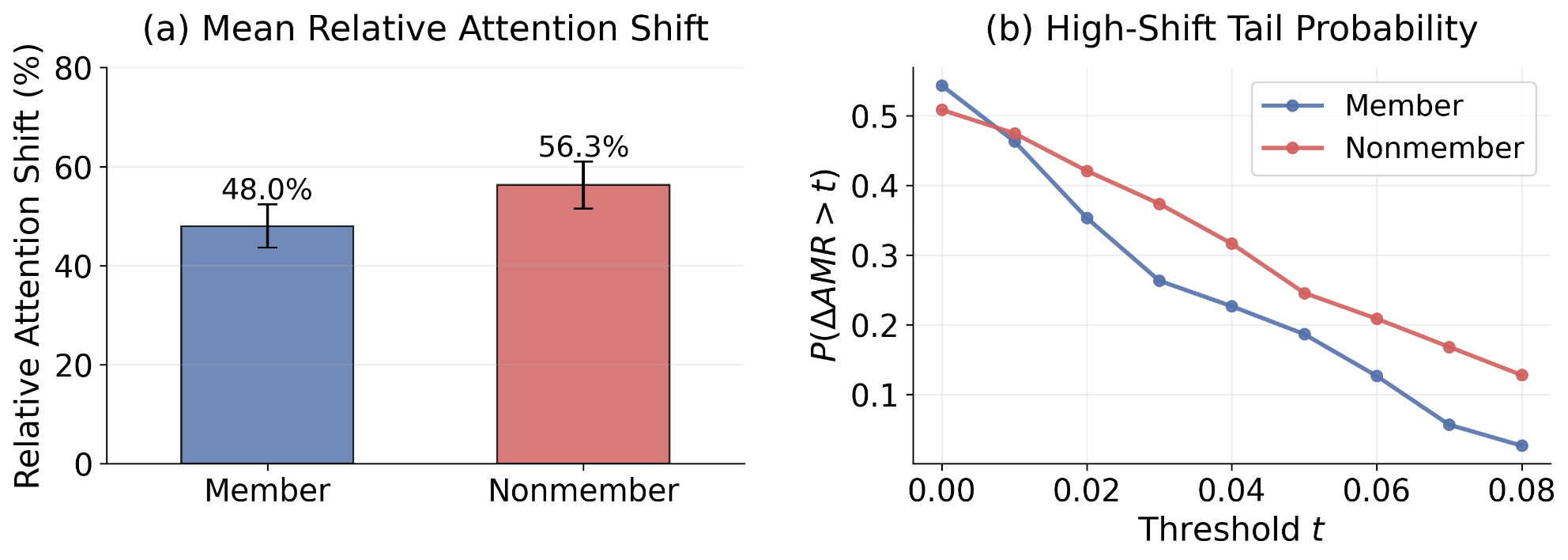}
    \caption{
    Attention shift under semantic distraction.
    (a) Mean relative attention shift for member and non-member samples.
    (b) Tail probability of large attention shifts, measured as $P(\Delta \mathrm{AMR} > t)$.
    %Non-member samples exhibit both higher average shift and heavier tails.
    Details are provided in Appendix~C.
    }
    \label{fig:attention_shift}
\end{figure}

\subsection{Mechanism Analysis}
\label{sec:mechanism}

The main results show that semantic distraction yields a stronger output-only membership signal than existing black-box probes. We further analyze why this signal emerges by examining how the inserted distractor changes visual attention during generation. The key intuition is that member samples remain anchored to the original image semantics, whereas non-member samples are more easily redirected toward the distractor. The attention analysis is diagnostic and not used by DistractMIA.

To quantify this effect, we measure the fraction of visual attention assigned to the image region where the distractor is inserted. For each sample, we compare this attention mass on the original image and on the distracted image, using the same spatial region in both cases. A larger shift means that the inserted distractor attracts more visual attention during generation. Figure~\ref{fig:attention_shift} reports the mean attention shift for member and non-member samples, together with the probability that the shift exceeds a threshold, which captures how often each group exhibits large attention reallocation.

The results support the semantic anchoring view. Non-member samples exhibit larger average attention shifts and heavier large-shift tails than member samples, indicating that their generation is more easily redirected toward the inserted region. This attention pattern provides a mechanistic explanation for the output-level signal used by DistractMIA: non-member responses are more likely to reflect distractor-induced changes, whereas member responses remain more stable.

% \subsection{Sensitivity Analysis}

% We further analyze two important factors in DistractMIA: the stability weight $\eta$ in the output-only distraction score and the number of stochastic generations $T$ used for each target image. The results are shown in Figure~\ref{fig:ablation}. Figure~\ref{fig:ablation}(a) shows the effect of different stability weights $\eta \in \{0.3,0.5,0.7,0.9\}$ on LLaVA. As $\eta$ increases from $0.3$ to $0.7$, performance consistently improves on both Flickr and DALL-E, indicating that generation stability provides an important membership signal. The best performance is achieved around $\eta=0.7$, while a further increase to $\eta=0.9$ slightly reduces performance. This suggests that relying excessively on stability weakens sensitivity to distractor-related semantic responses. Overall, balancing stability and distractor mention signals leads to the most effective membership inference performance. Figure~\ref{fig:ablation}(b) evaluates the effect of using different numbers of stochastic generations. Increasing $T$ from $3$ to $7$ gradually improves AUC, showing that additional generations help estimate behavioral consistency more reliably. At the same time, DistractMIA consistently outperforms KCMP and ModR\'enyi across all settings, demonstrating that the proposed behavioral signal remains effective even with a relatively small number of generations.

\subsection{Sensitivity Analysis}

\begin{figure}[t]
    \centering
    \includegraphics[width=\linewidth]{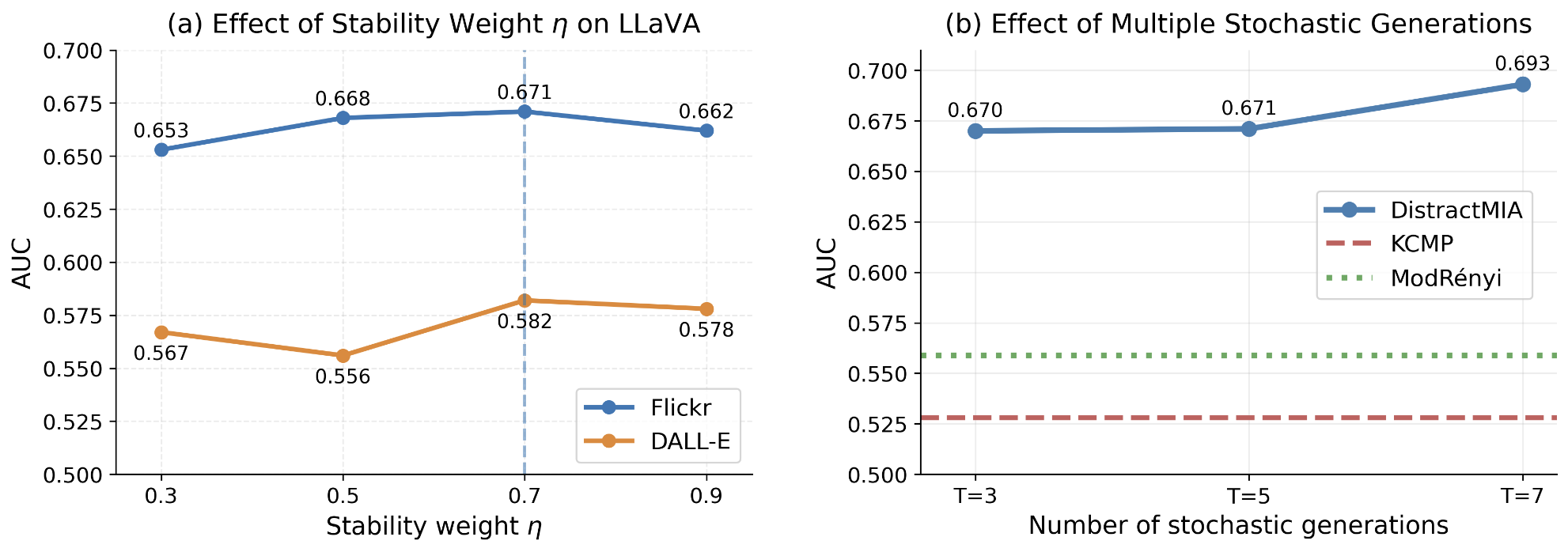}
    \caption{
    Ablation studies on DistractMIA.
    (a) Effect of stability weight $\eta$ in the output-only distraction score.
    (b) Effect of stochastic generation number $T$ used for behavioral estimation.
    }
    \label{fig:ablation}
\end{figure}

We analyze the stability weight $\eta$ in the output-only distraction score and the number of stochastic generations $T$ used to estimate each score. As shown in Figure~\ref{fig:ablation}(a), increasing $\eta$ from $0.3$ to $0.7$ improves performance on both Flickr and DALL-E, indicating that generation stability is an important membership signal. Performance slightly drops at $\eta=0.9$, suggesting that over-weighting stability reduces the contribution of distractor mentions. Figure~\ref{fig:ablation}Figure~\ref{fig:ablation}(b) shows that increasing $T$ from $3$ to $7$ steadily improves AUC, as more generations provide a more reliable estimate of response consistency. Across these choices of $T$, DistractMIA outperforms KCMP and ModR\'enyi, showing that the proposed signal remains effective with a small number of generations.

\subsection{Medical-Domain Extension}

Beyond mechanism analysis and sensitivity analysis, we further examine whether DistractMIA extends beyond natural-image benchmarks. We evaluate on the PMC medical-domain benchmark with LLaVA-Med, as shown in Table~\ref{tab:pmc_results}. Medical images often have lower semantic density than natural images, and clinically relevant information may be localized in subtle regions rather than object-level entities. This makes object-centric probing methods, such as KCMP, difficult to apply because clinically relevant cues in medical images are often subtle and do not always correspond to clear object boundaries.

DistractMIA remains effective in this setting. At the sample level, it achieves $0.692$ AUROC, clearly outperforming both the output-only Image Infer baseline and representative probability-access baselines. The advantage further grows with set-level aggregation, reaching $0.999$ AUROC at $K=30$, while the selected baselines improve only modestly. This suggests that semantic distraction remains effective even when images lack clear object-centric semantics.

\setlength{\abovecaptionskip}{6pt} 
\begin{table}[t]
\centering
\small
\setlength{\tabcolsep}{6pt}
\renewcommand{\arraystretch}{1.15}
\begin{tabular}{c|ccc|ccc}
\toprule
\multicolumn{7}{c}{\textbf{LLaVA-Med (PMC)}} \\
\midrule
$K$ & Perplexity & Min-K\% & MaxR\'enyi-K\% & Image Infer (R) & KCMP & \textbf{DistractMIA} \\
\midrule
1 & 0.523 & 0.498 & 0.545 & 0.486 & -- & \textbf{0.692} \\
10 & 0.524 & 0.517 & 0.565 & 0.512 & -- & \textbf{0.954} \\
30 & 0.556 & 0.607 & 0.604 & 0.428 & -- & \textbf{0.999} \\
\bottomrule
\end{tabular}
\vspace{2mm} 
\caption{Membership inference AUC on LLaVA-Med (PMC).}
\label{tab:pmc_results}
\end{table}

\section{Conclusion}

In this work, we study output-only black-box membership inference for vision-language models. We propose DistractMIA, a semantic-distraction framework that infers membership from response changes after inserting a controlled distractor. DistractMIA does not rely on logits, probabilities, or masked prediction correctness. Instead, it uses calibrated distractor selection and repeated textual responses to measure distraction-induced stability and semantic uptake. Experiments across multiple VLMs and benchmarks show that DistractMIA consistently outperforms output-only baselines and remains competitive with stronger-access methods. Further evaluation on a medical benchmark suggests that the same idea can extend to images with less explicit object-centric structure, while attention-shift analysis provides supporting evidence for the proposed distraction-based signal.

\paragraph{Limitations.}
DistractMIA relies on effective distractor selection and repeated generations, which may increase calibration and query costs. In addition, its current evaluation mainly focuses on single-turn image description settings. Extending semantic distraction to multi-turn or reasoning-intensive VLM tasks remains future work.

\bibliography{reference}
\bibliographystyle{unsrtnat}

%%%%%%%%%%%%%%%%%%%%%%%%%%%%%%%%%%%%%%%%%%%%%%%%%%%%%%%%%%%%
\appendix

\newpage

\section{Additional Method Details}
\label{app:method_details}

This section provides implementation details for the two stages of DistractMIA. Algorithm~\ref{alg:distractor_search} summarizes the reference-based search procedure used to select the semantic distractor configuration. Algorithm~\ref{alg:distractmia_inference} summarizes the output-only scoring procedure used during inference. The same optimized configuration is fixed before evaluation and applied to all audited samples.

\begin{algorithm}[t]
\caption{Semantic Distractor Configuration Search}
\label{alg:distractor_search}
\begin{algorithmic}[1]
\Require Target VLM $f$, calibration pool $\mathcal{C}$, candidate distractor space $\Omega$, prompt $q$, reward weights $\alpha, \beta, \lambda$
\Ensure Optimized distractor configuration $\mathcal{P}^*$

\State Initialize policy $\pi_\theta$ over configurations $\mathcal{P} \in \Omega$
\For{each search round}
    \State Sample a distractor configuration $\mathcal{P} \sim \pi_\theta$
    \State Apply $\mathcal{P}$ to all calibration images: $\tilde{\mathcal{C}} = \mathcal{T}(\mathcal{C}, \mathcal{P})$
    \State Partition reference set into two independent splits: $\mathcal{R}^{(1)}, \mathcal{R}^{(2)}$
    \For{$r \in \{1,2\}$}
        \State Query $f$ on perturbed images in $\mathcal{R}^{(r)}$
        \State Convert each response to normalized word set
        \State Compute distracted-response stability $S_{\mathrm{ref}}(x; \mathcal{P})$ for each image
        \State Compute member–non-member gap:
        \[
        \Delta_r(\mathcal{P}) = \mathbb{E}_{x \in \mathcal{R}_m^{(r)}} [S_{\mathrm{ref}}(x;\mathcal{P})] - \mathbb{E}_{x \in \mathcal{R}_n^{(r)}} [S_{\mathrm{ref}}(x;\mathcal{P})]
        \]
    \EndFor
    \State Compute reward:
    \[
    R(\mathcal{P}) = \alpha \max(0, \Delta_1(\mathcal{P})) + \beta \max(0, \Delta_2(\mathcal{P})) - \lambda |\Delta_1(\mathcal{P}) - \Delta_2(\mathcal{P})|
    \]
    \State Update policy $\pi_\theta$ using REINFORCE gradient:
    \[
    \nabla_\theta J(\theta) = \mathbb{E}_{\mathcal{P}\sim\pi_\theta} [(R(\mathcal{P}) - b_t) \nabla_\theta \log \pi_\theta(\mathcal{P})]
    \]
\EndFor
\State Select the best configuration:
\[
\mathcal{P}^* = \arg\max_{\mathcal{P}} R(\mathcal{P})
\]
\State \Return $\mathcal{P}^*$
\end{algorithmic}
\end{algorithm}

\begin{algorithm}[t]
\caption{Output-only Distraction Scoring}
\label{alg:distractmia_inference}
\begin{algorithmic}[1]
\Require Target VLM $f$, audited sample set $\mathcal{S} = \{X_i\}_{i=1}^K$, optimized configuration $\mathcal{P}^*$, prompt $q$, generation number $T$, distractor keywords $\mathcal{K}_{\mathrm{dist}}$
\Ensure Sample-level scores $A(X_i)$ and set-level score $A(\mathcal{S})$

\For{each sample $X_i \in \mathcal{S}$}
    \State Construct perturbed image $\tilde{X}_i = \mathcal{T}(X_i, \mathcal{P}^*)$
    \State Query $f$ on $X_i$ and $\tilde{X}_i$ for $T$ generations:
    \[
    \mathcal{Y}_{\mathrm{orig}} = \{y_1(X_i),\dots,y_T(X_i)\}, \quad
    \mathcal{Y}_{\mathrm{pert}} = \{y_1(\tilde{X}_i),\dots,y_T(\tilde{X}_i)\}
    \]
    \State Convert each response to normalized word sets $W_1,\dots,W_T$
    \State Compute response stability:
    \[
    S(X_i) = \frac{1}{\binom{T}{2}} \sum_{i<j} \frac{|W_i \cap W_j|}{|W_i \cup W_j|}, \quad
    S(\tilde{X}_i) \text{ analogously}
    \]
    \State Compute distractor mention rate:
    \[
    M(X_i) = \frac{1}{T} \sum_{i=1}^T \mathbf{1}[W_i \cap \mathcal{K}_{\mathrm{dist}} \neq \emptyset], \quad
    M(\tilde{X}_i) \text{ analogously}
    \]
    \State Compute distraction-response score:
    \[
    G(X_i) = \eta S(X_i) - (1-\eta) M(X_i), \quad
    G(\tilde{X}_i) \text{ analogously}
    \]
    \State Compute sample-level membership score:
    \[
    A(X_i) = G(\tilde{X}_i) - G(X_i)
    \]
\EndFor
\State Compute set-level score:
\[
A(\mathcal{S}) = \frac{1}{K} \sum_{i=1}^{K} A(X_i)
\]
\State \Return $A(X_i)$ and $A(\mathcal{S})$
\end{algorithmic}
\end{algorithm}

\section{Experimental Setup}
\subsection{Datasets}
\noindent \textbf{RL Calibration Data (COCO):}  
Used for the RL-based distractor configuration search in DistractMIA. The calibration set contains 100 images in total, including 50 member samples and 50 non-member samples. To compute the split-consistency reward in Section~3.3, these 100 images are randomly divided into two disjoint reference splits with a 60/40 ratio, corresponding to $\mathcal{R}^{(1)}$ and $\mathcal{R}^{(2)}$.

\noindent \textbf{VL-MIA/Flickr:}  
Contains 600 images (300 member + 300 non-member). Member images are sampled from Flickr images in MS COCO used by the target models, while non-member images are selected from photos uploaded after January 1, 2024. Following the official VL-MIA setup, we also use deliberately corrupted versions of member images to simulate more realistic auditing scenarios.

\noindent \textbf{VL-MIA/DALL$\cdot$E:}  
Used for semantically matched membership evaluation, with a total of 592 images. Member data are drawn from the intersection of the training images of LAION-CCS (including LAION, Conceptual Captions 3M/12M, and SBU captions), while non-member data are generated by DALL·E corresponding one-to-one to the member images. % This ensures a paired evaluation set for comprehensive MIA assessment.

\noindent \textbf{PMC-OA (Medical Data):}  
Used for evaluating medical VLLMs. Since the exact training data of LLaVA-Med is not fully available, PMC-OA is the only exception where the reference set and evaluation set come from the same dataset source. We strictly split PMC-OA into non-overlapping subsets: the reference subset is used only for distractor configuration selection, while the evaluation subset is used for final testing.

\subsection{Models}
We evaluate four core VLLMs covering both general and medical domains:

\noindent \textbf{LLaVA-1.5}: A general-domain autoregressive VLM with a Vicuna-7B decoder and a CLIP-ViT-L/14 visual encoder, totaling approximately 7B parameters. We use it for VL-MIA/Flickr and VL-MIA/DALL$\cdot$E evaluation.

\noindent \textbf{MiniGPT-4}: Multi-stage generative model, based on LLAMA2-7B with BLIP visual encoder features. Approximately 7B parameters, pre-trained on COCO and LAION-CCS, used for general image MIA evaluation.

\noindent \textbf{LLaMA2-Accessory}: Autoregressive model with modular visual encoder and LLAMA2-7B decoder, approximately 7B parameters, pre-trained on COCO and LAION-CCS, for general-domain evaluation.

\noindent \textbf{LLaVA-Med}: Medical VLLM with CLIP ViT-L/14 encoder and Vicuna-7B decoder, approximately 7B parameters, pre-trained on PMC-OA data for medical-domain evaluation.

\subsection{Computational Resources}

All experiments were conducted on a single NVIDIA L20 GPU with 40GB memory. Inference time per model is approximately 8 hours, varying depending on model size and dataset scale.

\section{Attention Shift Analysis under Semantic Distraction}
\label{sec:attention_shift}

To better understand the mechanism behind DistractMIA, we analyze how semantic distractors influence visual attention. The main question is whether inserting a distractor changes how the model attends to image regions, and whether this effect differs between member and non-member samples.

\paragraph{Quantifying attention shift.}
We first define the \emph{Attention Mass Ratio} (AMR) of the distractor region:
\begin{equation}
\mathrm{AMR}(x) =
\frac{\sum_{v \in \mathcal{P}} a_v}
{\sum_{v \in \mathcal{V}} a_v},
\end{equation}
where $\mathcal{V}$ denotes all visual tokens, $\mathcal{P}$ corresponds to tokens within the inserted distractor region, and $a_v$ represents the attention weight assigned to token $v$ during generation.

We then measure the attention shift caused by semantic perturbation:
\begin{equation}
\Delta \mathrm{AMR}(x) =
\mathrm{AMR}(x_{\mathrm{perturbed}}) -
\mathrm{AMR}(x_{\mathrm{original}}).
\end{equation}
To obtain a scale-invariant metric, we further normalize this shift as
\begin{equation}
\mathrm{Relative\ Shift}(x) =
\frac{\Delta \mathrm{AMR}(x)}
{\mathrm{AMR}(x_{\mathrm{original}})}.
\end{equation}
This quantity measures how strongly the model reallocates attention toward the distractor relative to its original attention structure.

\paragraph{Average and tail behavior.}
The results are shown in Figure~\ref{fig:attention_shift}.
In Figure~\ref{fig:attention_shift}(a), we report the average relative attention shift for member and non-member samples. Member samples exhibit an average shift of $48.0\%$, while non-members reach $56.3\%$. This indicates that non-member samples are more susceptible to semantic distraction, as their attention is more readily redirected toward the inserted region. In contrast, member samples maintain stronger focus on the original visual content.

Average differences alone do not fully characterize the membership signal. We therefore examine the tail distribution:
\begin{equation}
P(\Delta \mathrm{AMR} > t),
\end{equation}
which measures the probability of observing large attention shifts. As shown in Figure~\ref{fig:attention_shift}(b), the difference between members and non-members becomes more pronounced as the threshold $t$ increases. While both groups behave similarly at low thresholds, non-member samples consistently exhibit higher probabilities in the high-shift region, for example when $t \geq 0.02$. The gap further widens for larger $t$, suggesting that non-members are not only more affected on average, but also more likely to undergo large attention reallocation.

These observations are consistent with the semantic competition view. When a distractor is inserted, the model must balance the original visual semantics and the newly introduced distractor semantics. For member samples, the model has previously learned stable representations of similar inputs, so the original semantics tend to dominate the competition. For non-member samples, the absence of strong memorized representations makes the original semantics less dominant, allowing the inserted distractor to attract more attention.

This attention-level discrepancy explains the effectiveness of DistractMIA. Increased attention on the distractor region influences generation behavior, making outputs more likely to reference distractor-related semantics. These changes appear as observable signals, such as keyword mentions or behavioral shifts, which can be used for membership inference. The analysis suggests that membership information is not only reflected in output probabilities, but can also emerge from structured differences in how models respond to semantic perturbations.

\section{Case Study}

\begin{figure}[t]
    \centering
    \includegraphics[width=0.24\linewidth]{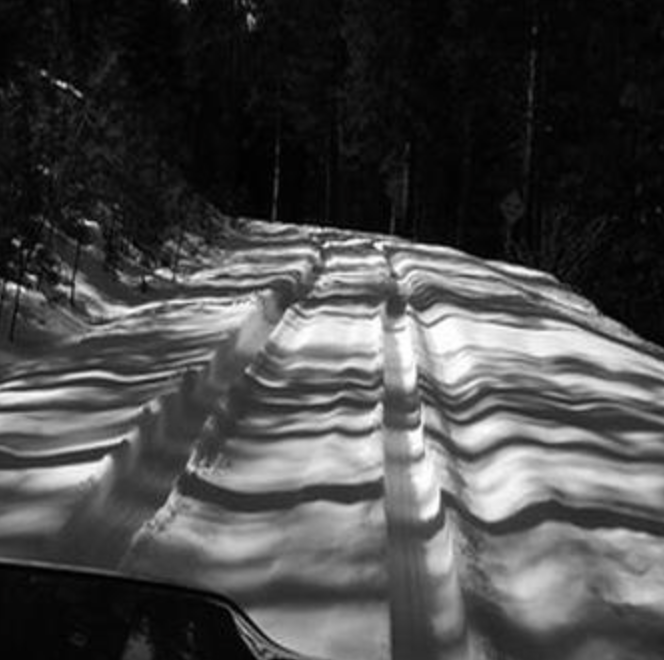}
    \includegraphics[width=0.24\linewidth]{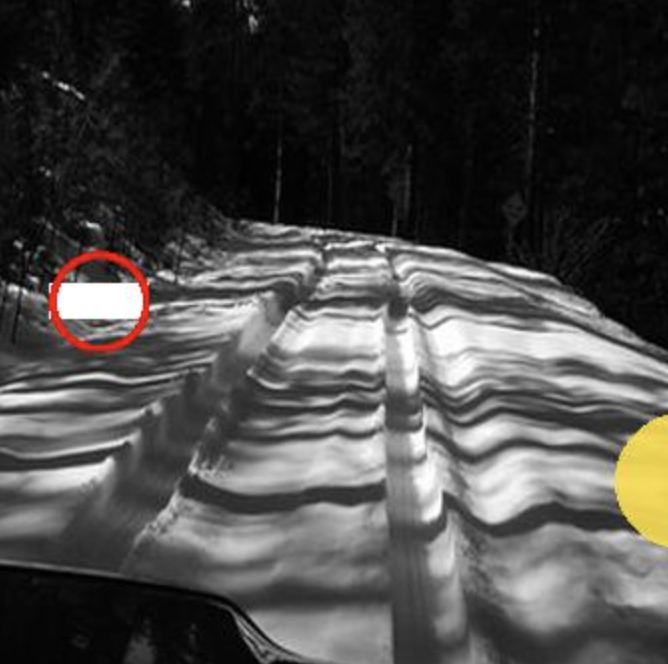}
    \includegraphics[width=0.24\linewidth]{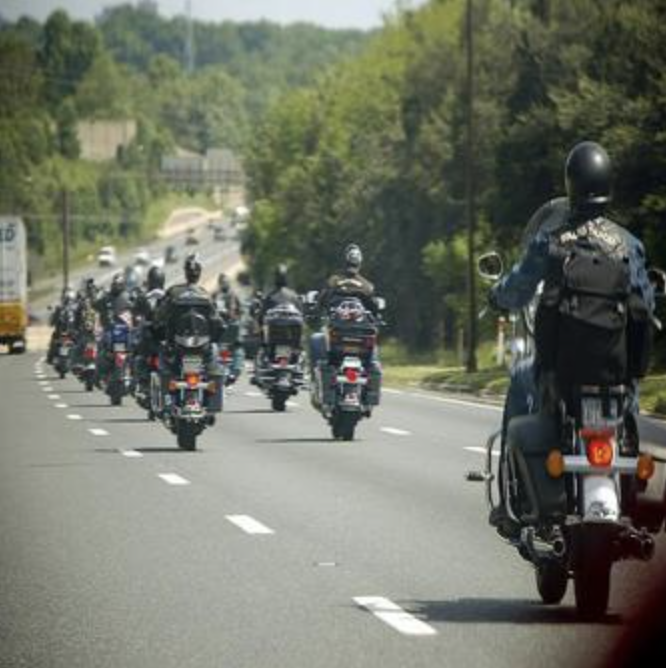}
    \includegraphics[width=0.24\linewidth]{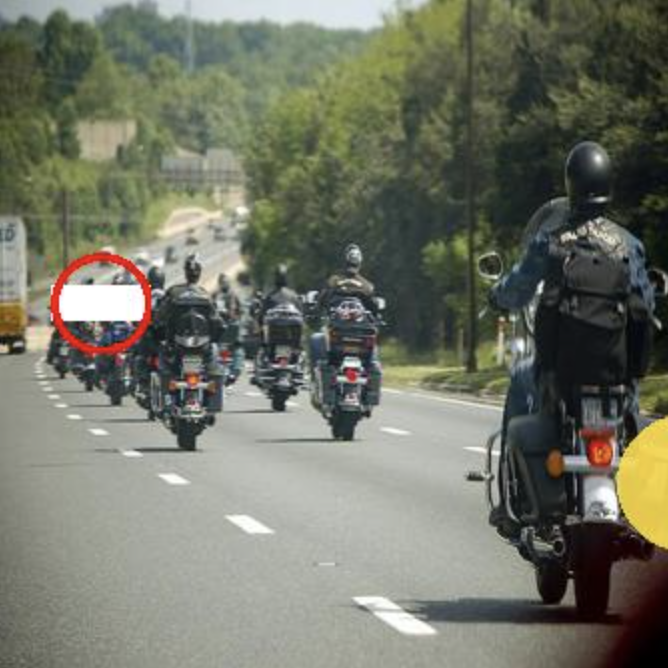}

    \caption{
    (a) Non-member (original), (b) Non-member (perturbed),
    (c) Member (original), (d) Member (perturbed).
    Non-member outputs are more influenced by distractors, while member outputs remain stable.
    }
    \label{fig:case_study}
\end{figure}

We present qualitative examples showing how semantic distractors influence model outputs differently for member and non-member samples. Figure~\ref{fig:case_study} compares original and perturbed inputs for both cases.

After applying the same distractor pattern, the two examples show different response behaviors. We first examine the non-member case, where the inserted distractor is explicitly incorporated into the generated description.

\paragraph{Non-member sample.}
\begin{itemize}[leftmargin=1.2em, itemsep=1pt, topsep=2pt]
\item \textbf{Original:} 
``The image captures a snowy road with a car driving down it... The snow-covered ground creates a serene and picturesque winter landscape.''

\item \textbf{Perturbed:} 
``The image depicts a snowy road with a \textbf{red circle} placed on the left side of the scene. The circle is likely a \textbf{warning or a sign}...''
\end{itemize}

The perturbed response not only mentions the ``red circle'' but also interprets it as a possible ``warning or sign,'' indicating that the generated description is redirected by the distractor.

\paragraph{Member sample.}
\begin{itemize}[leftmargin=1.2em, itemsep=1pt, topsep=2pt]
\item \textbf{Original:} 
``The image captures a group of motorcyclists riding down a highway...''

\item \textbf{Perturbed:} 
``The image captures a group of motorcyclists riding down a highway...''
\end{itemize}

In contrast, for the member sample, the perturbed response remains nearly unchanged and does not mention the inserted distractor. This contrast illustrates the output-level signal used by DistractMIA: non-member responses are more easily shifted by semantic distraction, whereas member responses remain more stable.

\newpage
\section*{NeurIPS Paper Checklist}

\begin{enumerate}

\item {\bf Claims}
    \item[] Question: Do the main claims made in the abstract and introduction accurately reflect the paper's contributions and scope?
    \item[] Answer: \answerYes{} % Replace by \answerYes{}, \answerNo{}, or \answerNA{}.
    \item[] Justification: The abstract and introduction clearly describe the main contributions of this paper: the proposal of DistractMIA, an output-only black-box membership inference framework for vision-language models based on semantic distraction. The claims regarding inserting known semantic distractors, measuring response changes, and capturing member vs. non-member differences without accessing logits or hidden states. The claims about applicability are supported by evaluations on both natural-image benchmarks and a medical-domain benchmark.

\item {\bf Limitations}
    \item[] Question: Does the paper discuss the limitations of the work performed by the authors?
    \item[] Answer: \answerYes{} % Replace by \answerYes{}, \answerNo{}, or \answerNA{}.
    \item[] Justification: The paper explicitly discusses several limitations of DistractMIA: it depends on effective distractor selection and repeated generations, which may increase calibration and query costs. Additionally, the current evaluation focuses on single-turn image description tasks, and extending the method to multi-turn or reasoning-intensive VLM tasks is identified as future work. These discussions are included in the Conclusion section.

\item {\bf Theory assumptions and proofs}
    \item[] Question: For each theoretical result, does the paper provide the full set of assumptions and a complete (and correct) proof?
    \item[] Answer: \answerNA{} % Replace by \answerYes{}, \answerNo{}, or \answerNA{}.
    \item[] Justification: The paper does not include formal theoretical results or proofs; it is primarily a method and experimental study.

\item {\bf Experimental result reproducibility}
    \item[] Question: Does the paper fully disclose all the information needed to reproduce the main experimental results of the paper to the extent that it affects the main claims and/or conclusions of the paper (regardless of whether the code and data are provided or not)?
    \item[] Answer: \answerYes{} % Replace by \answerYes{}, \answerNo{}, or \answerNA{}.
    \item[] Justification: The paper provides all necessary information to reproduce the main experimental results, including dataset sources, reference set usage, distractor configuration, repeated textual generations, and evaluation metrics across multiple VLMs and benchmarks.

\item {\bf Open access to data and code}
    \item[] Question: Does the paper provide open access to the data and code, with sufficient instructions to faithfully reproduce the main experimental results, as described in supplemental material?
    \item[] Answer: \answerNo{} % Replace by \answerYes{}, \answerNo{}, or \answerNA{}.
    \item[] Justification: While the code for DistractMIA is not publicly released, the paper provides detailed descriptions of the method, dataset usage (Flickr, DALL·E, PMC), distractor configuration, repeated textual generations, and evaluation metrics, allowing others to faithfully reproduce the experiments based on the paper. The code will be made publicly available upon acceptance.

\item {\bf Experimental setting/details}
    \item[] Question: Does the paper specify all the training and test details (e.g., data splits, hyperparameters, how they were chosen, type of optimizer) necessary to understand the results?
    \item[] Answer: \answerYes{} % Replace by \answerYes{}, \answerNo{}, or \answerNA{}.
    \item[] Justification: The paper provides all necessary experimental details to understand and reproduce the results, including dataset sources (Flickr, DALL·E, PMC), reference set usage, distractor configuration search, repeated textual generations, and output feature extraction methods, along with evaluation metrics across multiple VLMs and benchmarks.

\item {\bf Experiment statistical significance}
    \item[] Question: Does the paper report error bars suitably and correctly defined or other appropriate information about the statistical significance of the experiments?
    \item[] Answer: \answerNo{} % Replace by \answerYes{}, \answerNo{}, or \answerNA{}.
    \item[] Justification: The paper reports mean AUROC and set-level accuracy over multiple generations, but it does not include error bars, confidence intervals, or other formal statistical significance measures.

\item {\bf Experiments compute resources}
    \item[] Question: For each experiment, does the paper provide sufficient information on the computer resources (type of compute workers, memory, time of execution) needed to reproduce the experiments?
    \item[] Answer: \answerYes{} % Replace by \answerYes{}, \answerNo{}, or \answerNA{}.
    \item[] Justification: The paper specifies the use of GPUs (e.g., NVIDIA L20) and CUDA version for all experiments. While precise runtime is not provided, the hardware type and setup are described in sufficient detail to estimate the computational resources required for reproducing the experiments. These details are provided in Appendix~B.3.
    
\item {\bf Code of ethics}
    \item[] Question: Does the research conducted in the paper conform, in every respect, with the NeurIPS Code of Ethics \url{https://neurips.cc/public/EthicsGuidelines}?
    \item[] Answer: \answerYes{} % Replace by \answerYes{}, \answerNo{}, or \answerNA{}.
    \item[] Justification: The research uses only publicly available datasets and does not involve human subjects or sensitive personal data. The methodology conforms to the NeurIPS Code of Ethics in all respects.


\item {\bf Broader impacts}
    \item[] Question: Does the paper discuss both potential positive societal impacts and negative societal impacts of the work performed?
    \item[] Answer: \answerYes{} % Replace by \answerYes{}, \answerNo{}, or \answerNA{}.
    \item[] Justification: The paper discusses the dual-use nature of membership inference for VLMs. On the positive side, DistractMIA can support privacy auditing by helping data owners, model providers, and auditors assess whether sensitive or copyrighted images may have been used in training. On the negative side, the same technique could be misused to infer training-set membership for private data. The paper discusses these risks and frames the method as an auditing tool rather than a deployment-time attack.
    
\item {\bf Safeguards}
    \item[] Question: Does the paper describe safeguards that have been put in place for responsible release of data or models that have a high risk for misuse (e.g., pre-trained language models, image generators, or scraped datasets)?
    \item[] Answer: \answerNA{} % Replace by \answerYes{}, \answerNo{}, or \answerNA{}.
    \item[] Justification: The paper uses only publicly available datasets and does not involve high-risk or dual-use models. No additional safeguards for responsible release are necessary.

\item {\bf Licenses for existing assets}
    \item[] Question: Are the creators or original owners of assets (e.g., code, data, models), used in the paper, properly credited and are the license and terms of use explicitly mentioned and properly respected?
    \item[] Answer: \answerYes{} % Replace by \answerYes{}, \answerNo{}, or \answerNA{}.
    \item[] Justification: The paper cites all used public datasets and models (e.g., Flickr, DALL·E, PMC) and follows their respective license and usage terms.

\item {\bf New assets}
    \item[] Question: Are new assets introduced in the paper well documented and is the documentation provided alongside the assets?
    \item[] Answer: \answerNA{} % Replace by \answerYes{}, \answerNo{}, or \answerNA{}.
    \item[] Justification: The paper does not release any new datasets, code, or models; all resources used are publicly available.

\item {\bf Crowdsourcing and research with human subjects}
    \item[] Question: For crowdsourcing experiments and research with human subjects, does the paper include the full text of instructions given to participants and screenshots, if applicable, as well as details about compensation (if any)? 
    \item[] Answer: \answerNA{} % Replace by \answerYes{}, \answerNo{}, or \answerNA{}.
    \item[] Justification: The paper does not involve crowdsourcing or human subjects research.

\item {\bf Institutional review board (IRB) approvals or equivalent for research with human subjects}
    \item[] Question: Does the paper describe potential risks incurred by study participants, whether such risks were disclosed to the subjects, and whether Institutional Review Board (IRB) approvals (or an equivalent approval/review based on the requirements of your country or institution) were obtained?
    \item[] Answer: \answerNA{} % Replace by \answerYes{}, \answerNo{}, or \answerNA{}.
    \item[] Justification: The paper does not involve human subjects research and therefore does not require IRB approval.

\item {\bf Declaration of LLM usage}
    \item[] Question: Does the paper describe the usage of LLMs if it is an important, original, or non-standard component of the core methods in this research? Note that if the LLM is used only for writing, editing, or formatting purposes and does \emph{not} impact the core methodology, scientific rigor, or originality of the research, declaration is not required.
    %this research? 
    \item[] Answer: \answerNA{}. % Replace by \answerYes{}, \answerNo{}, or \answerNA{}.
    \item[] Justification: LLMs were used only for writing, editing, and formatting assistance. They were not used as an important, original, or non-standard component of the core method, experiments, analysis, or scientific claims.

\end{enumerate}
\end{document}